\newif\ifNIPS

\NIPSfalse
\NIPStrue

\ifNIPS
   \documentclass{article}
   \usepackage{nips13submit_e,times}
\else
   \documentclass[conference,11pt]{IEEEtran}
   \IEEEoverridecommandlockouts

\fi

\usepackage[T1]{fontenc}
\usepackage{array}
\usepackage{multirow}
\usepackage{hhline}
\usepackage{hyperref}
\usepackage{color}
\usepackage{url}
\usepackage[pdftex]{graphicx}

\nipsfinalcopy

\begin{document}
\title{Curriculum Learning for Handwritten Text Line Recognition}

\ifNIPS
   \author{
      J\'er\^ome Louradour, Christopher Kermorvant \\
      A2iA S.A.\\
      39 rue de la Bienfaisance\\Paris 75008 France\\
      \texttt{\{jl,ck\}@a2ia.com}
   }
\else
   \author{
      \IEEEauthorblockN{J\'er\^ome Louradour, Christopher Kermorvant}
      \IEEEauthorblockA{A2iA S.A.	\\
      39 rue de la Bienfaisance\\Paris 75008 France\\
      \{jl,ck\}@a2ia.com
      }
   }
\fi

\maketitle

\begin{abstract}
Recurrent Neural Networks (RNN) have recently achieved the best performance in off-line Handwriting Text Recognition.
At the same time, learning RNN by gradient descent
leads to slow convergence, 
and training times are particularly long
when the training database consists of full lines of text.
In this paper, we propose an easy way
to accelerate stochastic gradient descent in this set-up,
and in the general context of learning to recognize sequences.
The principle is called Curriculum Learning, or shaping.
The idea is to first learn to recognize short sequences before training on all available training sequences.
Experiments on three different handwritten text databases (Rimes, IAM, OpenHaRT) show that
a simple implementation of this strategy
can significantly speed up the training of RNN for Text Recognition,
and even significantly improve performance in some cases.

\end{abstract}

\ifNIPS
\else

   \IEEEpeerreviewmaketitle
\fi

\section{Introduction}

The application of interest in this paper is off-line Handwritten Text Recognition (HTR),
on images of paper documents.
At the time being,
the most powerful models for this task
are Recurrent Neural Networks (RNN)
with
several layers of
multi-directional Long-Short Term Memory (LSTM) units~\cite{Graves2008RnnHandwriting,Menasi2012}.
Gradient-based optimization
of RNN, which cannot be guaranteed to converge to the optimal solution,
is a particularly hard issue for two reasons:

First, if we conceptually unfold the recurrences done in the spatial domain (2D, sometimes 1D),
we can see RNN as deep models.
Because of the numerous non-linear functions that are composed,
they are exposed to the burden of exploding and vanishing gradient~\cite{Bengio94VanishingGradient,Hochreiter1998RnnDifficulty,Pascanu2012RnnDifficulty}.
In practice, the use of LSTM units, which are carefully designed cells with multiplicative gates to store information over long periods and forget when needed,
turned out to be 
a key ingredient to enable the learning of RNN with standard gradient descent despite the network deepness.
There are other ways to efficiently learn RNN,
namely enhanced optimization approaches such as second-order methods~\cite{Martens2011RNN_HF}
or good initialization with momentum~\cite{Sutskever2013DNNMomentum}.
These methods are beyond the scope of this paper.

Secondly, RNN are used here for an unconstrained ``Temporal Classification'' task~\cite{Graves2006RNN_CTC,Graves2009RNN_CTC},
where the length of the sequence to recognize is in general different from the length of the input sequence.
In HTR, the goal is to detect occurrences of characters within a stream of image,
without {\it a priori} segmentation, in other words without knowing the alignment between the pixels and the target characters.
So the models must be optimized to solve two problems at the same time:
localizing the characters \emph{and}
classifying them.

Because of all these aspects,
training RNN takes a particularly long time.
Here we propose to make the training process more effective by using
the concept of Curriculum Learning,
that has already been successfully applied in the context of deep models and Stochastic Gradient Descent%
~\cite{CurriculumLearningBengio2009}.
The key idea is to guide the training by carefully choosing samples
so as to start simple and progressively increase the complexity of training samples.
The main motivation is to speed up the learning progression, without any loss of generality in the end.
Gradually increasing the complexity of the task
has been demonstrated to make learning faster and more robust in several scenarios.
This idea has been
exploited
in classification~\cite{CurriculumLearningBengio2009},
grammar induction~\cite{CurriculumGrammarElman93,CurriculumGrammarSpitkovsky, CurriculumGrammarKewei},
robotics~\cite{CurriculumRoboticSander},
cognitive science~\cite{CurriculumCognitiveKrueger}
and human teaching~\cite{CurriculumHumanTeachKhan}.

In this paper, we show how the Curriculum Learning concept can be naturally be applied to RNN
in the context of Handwritten Text Recognition,
using the text sequence length as a measure of its complexity.
We give empirical evidences that our proposal significantly
speeds up the learning progression.
The principle is general enough to be applied to any sequence recognition task,
and to any kind of model optimized using a gradient-based method.

\section{A curriculum for Text Recognition}

\subsection{Two tasks when learning to recognize text: Localization and Classification}

In text recognition,
locating the characters is necessary to learn to recognize them.
However, in many public database for Handwriting Recognition,
the positions and the text content are given for each page or paragraph, not for characters.
The localization of the lines is reasonably easy to obtain using automatic line segmentation.
But locating the characters is a more difficult problem,
particularly in the case of handwritten text where even humans can disagree on how to segment the characters.
This is why a Connectionist Temporal Classification (CTC) approach as proposed by~\cite{Graves2006RNN_CTC}
is a very practical way to train RNN models without intensive labeling effort.

In their CTC approach, {\cite{Graves2006RNN_CTC}
efficiently compute and derive a cost function that is the Negative Log-Likelihood (NLL),
with the assumptions that
all the frame probabilities are independent
and that all possible alignments are equiprobable.
Besides, an additional label is considered:
the {\it blank}, which stands for ``no character'' but also for ``zone in-between two characters'',
meaning that the {\it blank} label can be produced between any two different characters.
To the best of  our knowledge, taking into account all the possible alignments (including the {\it blank}) is the most effective approach in training RNN to detect characters
.
But it also unveils a vague localization of the characters,
especially at the beginning of the training process, 
when the RNN gives quasi-random guesses for the posteriors of the labels
(see the CTC Error Signal of Figure 4 in~\cite{Graves2006RNN_CTC}).

Several studies about Text Recognition
have revealed that the training process of RNN is particularly long~\cite{Schambach2013}.
Not only because of the heavy computational complexity due to the recurrences,
but also because the learning progression frequently starts with a plateau.
A high number of model parameter updates is needed before the cost function starts to decrease.
In some extreme cases, the learning seems to never start, as if the optimization process quickly got stuck in a poor local minimum.

\subsection{Building a suitable Curriculum}

One of the reasons for this difficulty to start learning is
the fact that when initializing with quasi-random model parameters,
the RNN has little chance to produce a reasonable segmentation.
Moreover, it is clear that the longer the sequences
are, the more serious the problem is.
In a nutshell, it is hard and inefficient to learn long sequences at first.

Thinking in the same spirit as~\cite{CurriculumLearningBengio2009},
let us make a parallel with how to teach kids to read and write.
A natural way is to do it step by step: first teach him to recognize characters by showing him isolated symbols,
then teach him short words, before introducing longer words and sentences.
A similar Curriculum Learning procedure can be done when optimizing neural networks by gradient-descent ({\it e.g.} RNN using CTC):
First optimize on a database of isolated characters (if available),
then on a database of isolated words,
and finally on a database of lines\footnote%
{RNN cannot decode paragraphs, just single lines:
the common RNN architectures collapse the 2D input image into a 1D signal just before aligning using CTC~\cite{Graves2008RnnHandwriting}.}.

Since having access to the positions of characters and/or words
may be costly or impossible,
we propose here to adapt this proposition to the case where only lines can be robustly extracted from the training database.
Keeping in mind that the difficulty when starting to train RNN
is related to the length of the training sequences,
a general way to build a Curriculum Learning for Text Line Recognition
is to first train on short lines, before including long ones.
Note that the last line in a paragraph can sometimes consist of a single word.

\subsection{Proposal: continuous curriculum}
\label{ssec:continuous_curri}

\def\idxSample{t}
\def\SampleInput{\mathbf{X}_\idxSample}
\def\SampleTarget{Y_\idxSample}
\def\SampleReco{\widehat{f}(\SampleInput)}
\def\CurriHyperParam{\lambda}
\def\NormFactor{{N_\CurriHyperParam}}

In practice,
it is awkward to build a step-wise schedule by splitting a database with respect to the sequence lengths.
Instead, we prefer to handle a probability to draw a sample line from the training database.
The idea of defining such a probability for probing the training database has already been successfully applied in Active Learning%
~\cite{Saar04ActiveSampling,Borisov2011ActiveLearning}. 

If $(\SampleInput,\SampleTarget)$ denotes a training sample (an image along with the corresponding target sequence of labels),
we propose to draw this sample with the following probability parameterized by~$\CurriHyperParam$:
\begin{equation}
\label{eq:proba_sampling}
P_\CurriHyperParam\left(\mbox{train on }(\SampleInput,\SampleTarget)\right) = \frac{1}{\NormFactor} \Big(shortness(\SampleInput,\SampleTarget)\Big)^{\CurriHyperParam}
\end{equation}
where
\begin{itemize}
\item $\NormFactor= \sum_\idxSample \left(shortness(\SampleInput,\SampleTarget)\right)^{\CurriHyperParam}$ is a normalization constant
      so that (\ref{eq:proba_sampling}) defines a probability over the set of all the available training samples.
     
\item $shortness\in[0,1]$ is a bounded value to represent how easy is a training sample.
      Here it is based on the sequence length.
      We discuss this quantity below.
\item $\CurriHyperParam\geq0$ is an hyper-parameter to tune how much the short words are favoured.
\end{itemize}
The particular setting $\CurriHyperParam=0$ amounts to the baseline approach where samples are drawn randomly with flat probability and with replacement.
So $\CurriHyperParam$ can be tuned during the training process.
In our experiments, we start with $\CurriHyperParam=3$
and linearly decrease $\CurriHyperParam$ until 0, during the equivalent of the first 5 epochs of training.
And one epoch is about 10k to 100k different lines of text (see number of labelled lines in table~\ref{tab:datasets}).

Concerning the $shortness$ measure,
we propose to use the following simple form:
\begin{equation}
\label{eq:shortness}
shortness(\SampleInput,\SampleTarget) = \frac{1}{\max(m,|\SampleTarget|)}
\end{equation}
where
$|\SampleTarget|$  is the length of the target sequence (number of characters),
and
$m>0$ is a minimal length which stands as a clipping threshold.
      Using $m=1$ is needed to avoid numerical problems when there are empty target sequences in the training set.
      Using more than~$1$ can be useful to avoid favouring too much on very small words, such as frequent pronouns,
      or punctuation marks when they are considered as a word with a single character.
      We used $m=5$ in our experiments, as it is a common length for short words.

Note that we could use in~(\ref{eq:shortness}) the width of the input images $|\SampleInput|$
instead of (or along with) the sequence length~$|\SampleTarget|$.
It also makes sense
and these two measures are actually correlated.
But the target length $|\SampleTarget|$
has the advantage of being independent of the resolution of the images,
and is also a notion that can be used in other applications than Vision.

\section{Experiments}

\subsection{Databases}

Three notated public handwriting datasets are used to evaluate our system:
\begin{itemize}
\item the IAM database, a dataset containing pages of handwritten English text~\cite{IAM},
\item the Rimes database\footnote{\url{http://www.rimes-database.fr}}, a dataset of handwritten French letters used in several ICDAR competitions
(lastly ICDAR 2011~\cite{ICDAR2011,Menasi2012}).
\item The OpenHaRT database, a dataset of handwritten Arabic pages,
used in two NIST Open Handwriting Recognition and Translation Evaluation
(lastly OpenHaRT 2013).
\end{itemize}
For all these databases, the localization of the words is available.
So we could compare the continuous Curriculum strategy proposed in section~\ref{ssec:continuous_curri}
with the simple ``by-hand'' Curriculum which consists in first training RNN to recognize words
and secondly to recognize lines.

We use distinct subsets of pages to train and to evaluate RNN models.
In the case of the ``by-hand'' Curriculum,
we carefully used the same subset of pages in the training set
of words
and in the training set
of lines.
Table~\ref{tab:datasets} gives the number of data in each training set.
\def\NOVLINE{\multicolumn{1}}
\begin{table}[h]
\begin{centering}
\begin{tabular}{|ll|c||rr||r|}
\hhline{~~~|---|}
   \multicolumn{3}{c|}{}              & \multicolumn{3}{c|}{ Training subset } \\
\hhline{--|-||~~~|}
         & \NOVLINE{l}{} & \# different   & \# labelled & \NOVLINE{r}{\# characters} & \# labeled \\
Database & \NOVLINE{l}{Language} & characters (*) & lines      &    \NOVLINE{r}{(in lines)} &    words   \\
\hhline{:==:=::==t::=:}
IAM    &  English  &        78 & 9\,462 &    338\,904 &  80\,505  \\
\hhline{:==:=::==::=:}
Rimes  &    French &       114 & 11\,065 &    429\,099 & 213\,064  \\
\hhline{:==:=::==::=:}
OpenHaRT &    Arabic       & 154 & 91\,811 & 2\,267\,450 & 524\,196 \\ 
\hhline{--|-||--||-|}
\end{tabular}
\par\end{centering}
\caption{%
Number of data in the training sets used in this paper.
\newline
(*) The number of different characters depends on the language (for instance there are some diacritics in French that do not exist in English)
and also on the punctuation marks that have been labelled in the database.
All tasks are case-sensitive.
For Arabic recognition, we used a fribidi conversion that map 37 Arabic symbols into 128 different shapes.
}
\label{tab:datasets}
\end{table}

The resolution of images to feed the network is fixed to 300\,dpi.
Original OpenHaRT images ({\it resp.} Rimes images) are in 600\,dpi ({\it resp.} 200 dpi) and they were rescaled with a factor 0.5 ({\it resp.} 1.5), using interpolation.

\subsection{Modeling and learning details}

\begin{figure*}[hb]
\centering
\includegraphics[width=\textwidth]{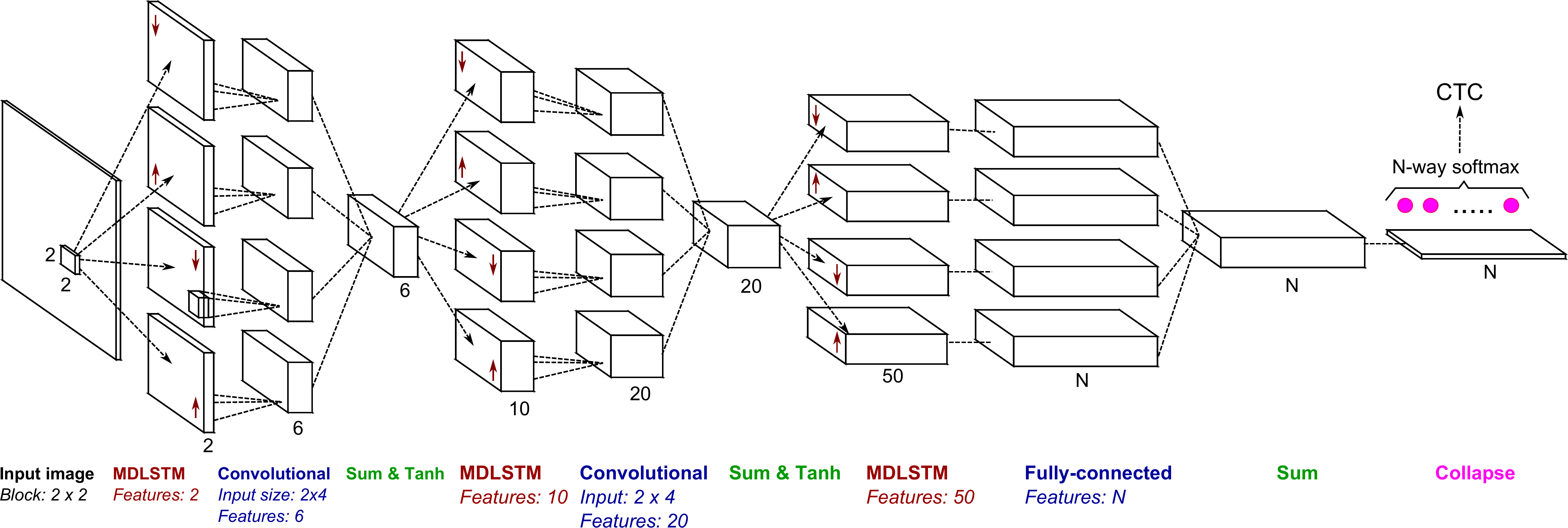}
\caption{%
   RNN topology used for all the experiments of this paper.
   The resolution of input images is 300\,dpi.
   The size of the hidden layers is given as the number of features in each intermediate representation.
   $N$~is the number of possible target characters including the {\it blank} (``\# different characters'' in Table~\ref{tab:datasets}, plus one).
}
\label{fig:RNNarchi}
\end{figure*}
The RNN topology we use is depicted in figure~\ref{fig:RNNarchi}.
It is the same as described in~\cite{Graves2008RnnHandwriting},
except that the sizes of the filters have been adapted to images in 300\,dpi:
we used 2x2 input tiling,
and 2x4 filters in the two sub-sampling hidden layers (which are convolutional layers without overlap between the filters).
The LSTM layers scan the inputs in 4 directions,
and the computations can be parallelized over the 4 directions.

All the models were optimized using Stochastic Gradient Descent~\cite{SGDLecun98}:
a model update happens after each training sample ({\it i.e.} each line of characters) is visited.
The learning rate is constant, and was fixed to $0.001$ in all our experiments.

\subsection{Performance assessment}

As we are interested in convergence speed,
we plot convergence curves that represent the evolution of some costs
with respect to a unit of progression of the training algorithm.
In our case, we use Stochastic Gradient Descent~\cite{SGDLecun98}
and the unit of progression could be for instance
the number of updates, that is the number of training samples that have been browsed.
Given that the sequence length of training samples is the measure of complexity
here,
we chose instead to represent the progression by the total number of targets (characters) that have been browsed.
This unit is more representative of the computation time than the number of updates,
because the inputs are sequences with variable-length,
and we remind that the Curriculum strategies tend to process shorter sequences in the beginning of the learning process.

We remind that the cost optimized using CTC~\cite{Graves2006RNN_CTC} is the
Negative Log-Likelihood~(NLL), which can be averaged over the number of training sequences.
However, probabilities decrease exponentially with sequence length.
For this reason, the NLL average costs are usually higher on databases with long sequences ({\it e.g.} lines) than on databases with short or middle-length sequences ({\it e.g.} words).
That is why we chose a normalized NLL to monitor the performance of our systems:
\begin{equation}
\label{eq:normNLL}
\mbox{normNLL}\left( \big\{(\SampleInput,\SampleTarget)\big\} \right)= \frac{\sum_\idxSample \mbox{NLL}(\SampleTarget|\SampleInput)}{\sum_\idxSample |\SampleTarget|}
\end{equation}

As a relevant but discrete cost to evaluate RNN optical models,
we also monitor the Character Error Rate (CER) that is computed by an edit distance, normalized in a similar manner:
\begin{equation}
\label{eq:CER}
\mbox{CER}= \frac{\sum_\idxSample EditDistance(\SampleTarget,\SampleReco)}{\sum_\idxSample |\SampleTarget|}
\end{equation}
where $\SampleReco$ is the most probable sequence recognized by the RNN.
The edit distance is a \emph{Levenshtein distance} defined by the minimum
number of insertions, substitutions and deletions required to change the target~$\SampleTarget$ into the model's prediction $\SampleReco$.

\subsection{Results and analysis}

\ifNIPS
\def\figwidth{0.75\columnwidth}
\else
\def\figwidth{0.85\columnwidth}
\fi

\begin{figure}[ht]
\centering
\includegraphics[width=\figwidth]{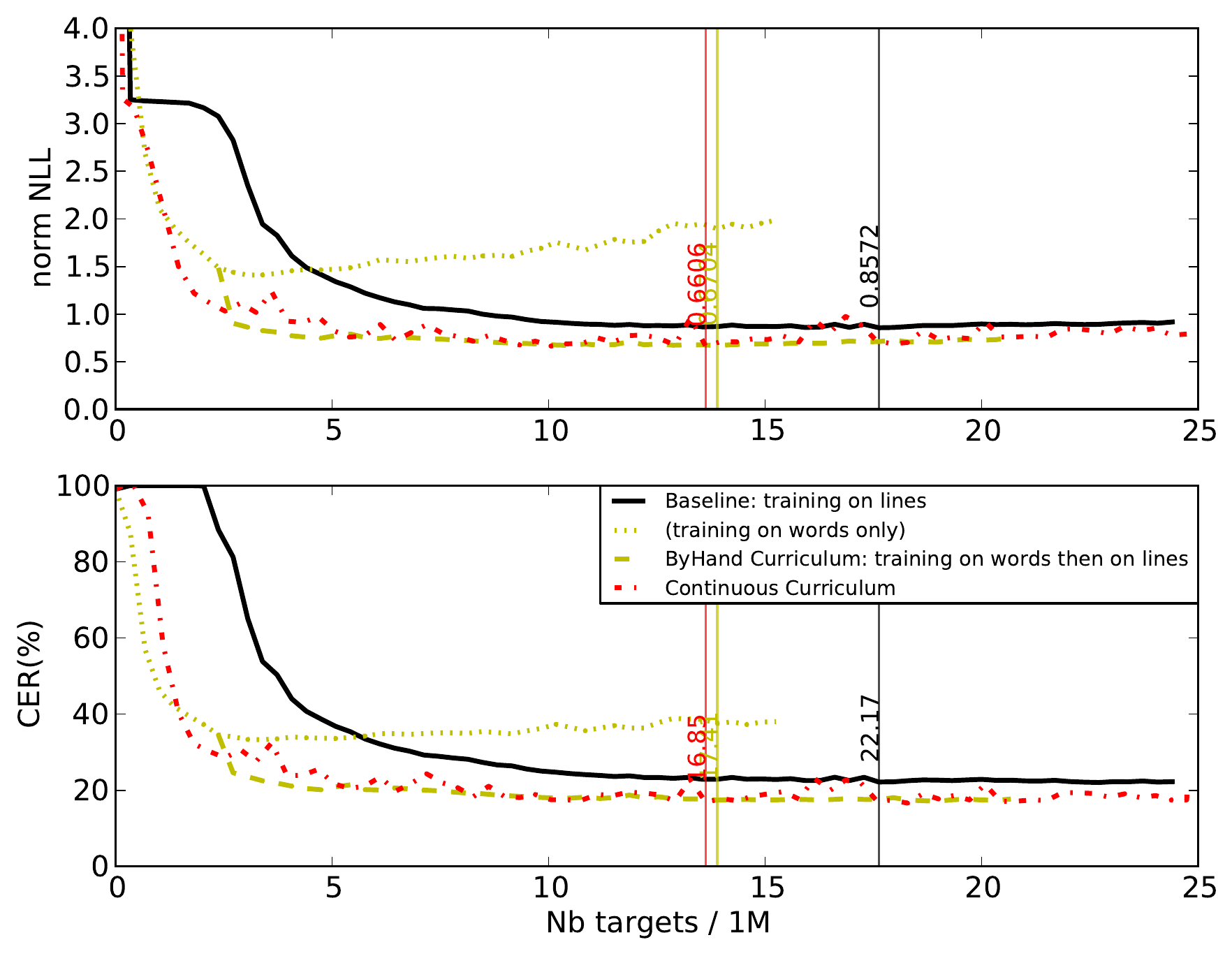}
\caption{Convergence Curves for English Handwritten Text Recognition (IAM).}
\label{fig:resultEnglish}
\end{figure}

\begin{figure}
\centering
\includegraphics[width=\figwidth]{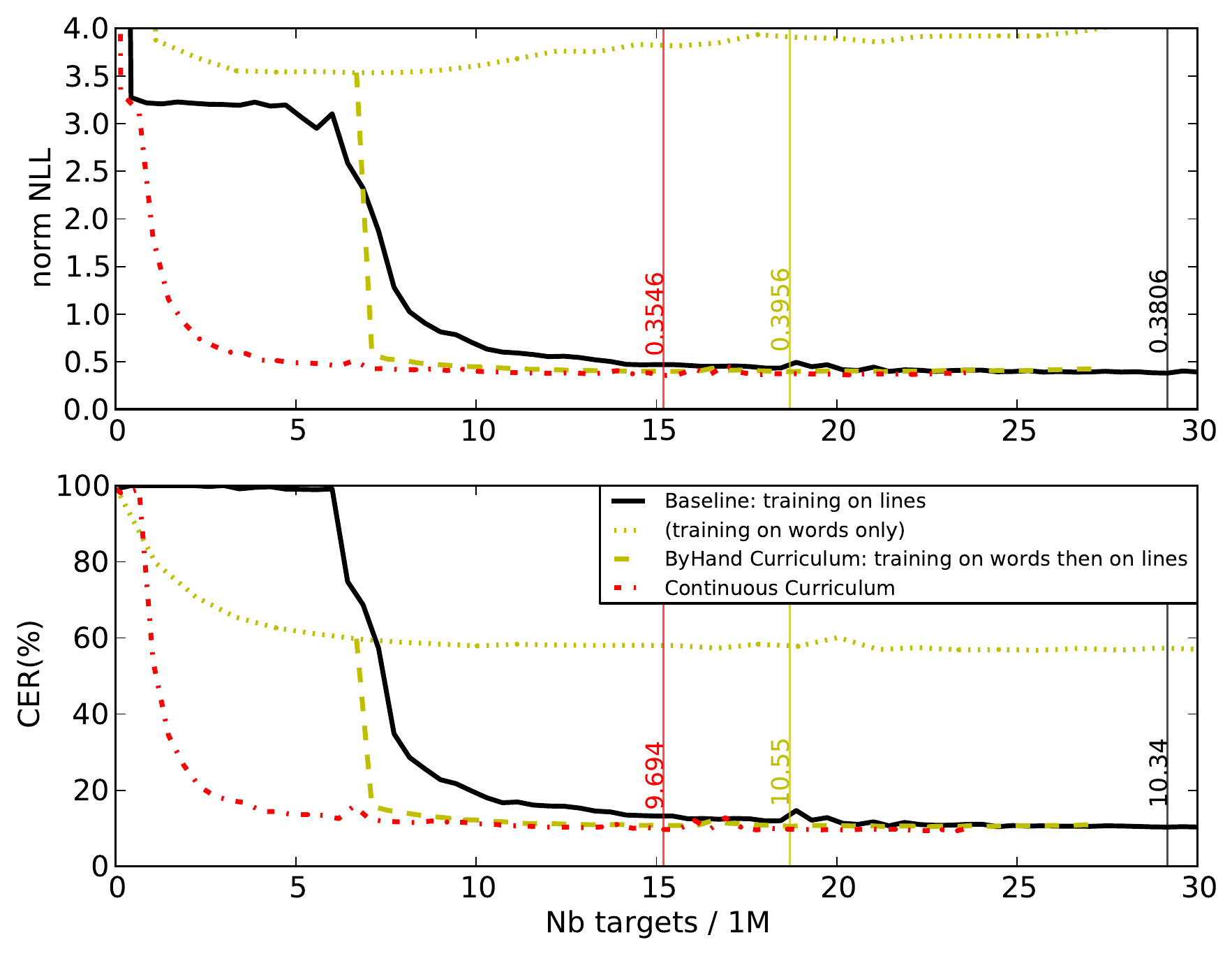}
\caption{Convergence Curves for French Handwritten Text Recognition (Rimes).}
\label{fig:resultFrench}
\end{figure}

\begin{figure}
\centering
\includegraphics[width=\figwidth]{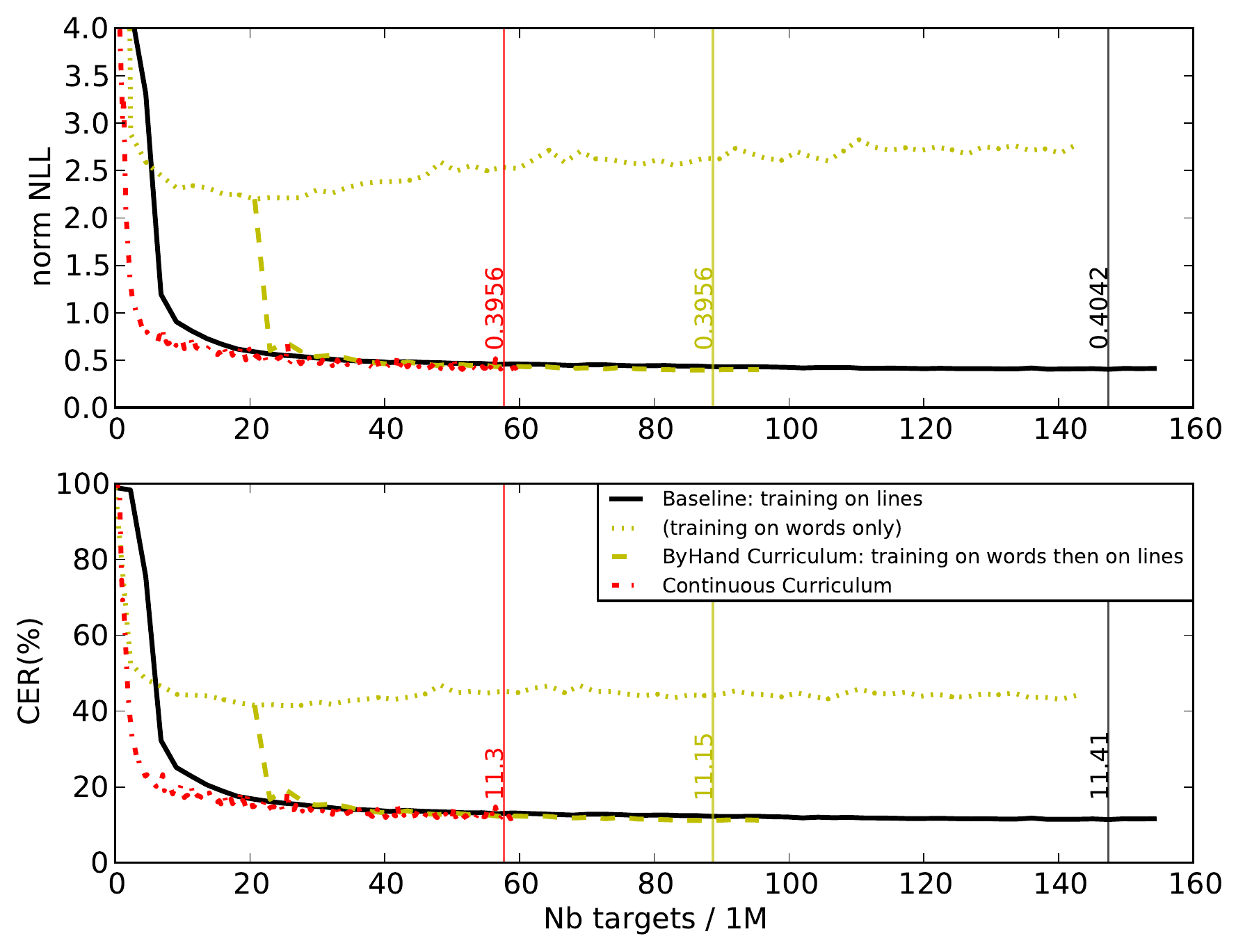}
\caption{Convergence Curves for Arabic Handwritten Text Recognition (OpenHaRT).}
\label{fig:resultArabic}
\end{figure}

The convergence curves for learning Handwritten Text Recognition on the three languages,
respectively IAM (English), Rimes (French) and OpenHaRT (Arabic),
are shown in Figures~\ref{fig:resultEnglish},~\ref{fig:resultFrench}~and~\ref{fig:resultArabic}.
They show the progression of the costs on the validation dataset,
and the vertical lines point out the best cost values achieved during all the learning process.

All the systems use exactly the same RNN topology and the same optimization procedure,
the only difference is the way the training samples are drawn.
The baseline system, represented by the black solid curves, consists in shuffling the dataset differently at each epoch,
{\it i.e.} randomly drawing training samples with flat priors and without replacement.
The dashed and dotted yellow curves represent the convergence obtained with a ``by-hand'' curriculum,
starting to train on isolated words, and then training on lines
(the switch was done when no more improvement is made by continuing training of words, looking at the performance on the validation set of lines).
Finally, the red dashed curve represents the continuous Curriculum approach presented in section~\ref{ssec:continuous_curri}.

In the case of the IAM database, a great improvement is achieved by using Curriculum Learning, without any additional training time:
the CER\% is decreased from 22\% to about 17\%.
In the case of the Rimes and the OpenHaRT databases, the improvement in performance is slight,
but the rate of convergence speed up is remarkable:
the whole learning process is roughly twice shorter.

The impact of the Curriculum strategy is visible at the beginning of the training,
where the cost functions can be decreased very fast.
However, after this initial fast training phase has been completed, and after a transitory phase,
the convergence rates are suddenly particularly low whereas the training is not finished.
Yet this difficulty to ``stop learning'' affects all the systems,
and indicates that another strategy than the Curriculum should be used to speed up this last learning phase.
For instance, techniques to compute a forced alignment~\cite{Schambach2013}.

Additional experiments show that, when \emph{adapting} a RNN that has already been trained and that is able to recognize a good part of the characters,
the Curriculum strategy does \emph{not} improve over the purely random baseline strategy (neither in performance nor in speed),
even in cases where the CER\% was high on the new database on which to adapt.
This confirms that implementing a Curriculum based on the sequence length
can play a crucial role at the beginning of the learning process,
but does not affect convergence speed any more  after the RNN has learned to detect the positions of the characters.

The fact that Curriculum Learning can improve generalization performance supports one point
mentioned by~\cite{Erhan2010UnsupervisedPreTraining},
namely that the networks optimized by stochastic gradient descent are greatly influenced by early training samples.
By choosing these samples and modifying the initial learning steps,
Curriculum learning is similar to other methods devoted to optimize deep models
such as careful initialization~\cite{Sutskever2013DNNMomentum}
and unsupervised pre-training~\cite{Bengio2006DBN,Erhan2010UnsupervisedPreTraining}.
However, it is complementary and can be used in combination with these methods.

\section{Conclusion}

This paper describes an easy-to-implement strategy to speed up the learning process,
that can also provide better performance in the end.
The principle is to build a curriculum based on the lengths of the target sequences.
Experimental results show that in the case of Recurrent Neural Network for text line recognition optimized by stochastic gradient descent,
the first phase of the learning can be drastically shorten,
and the generalization performance can be improved, especially when the training set is limited.

At the same time, the slowness of the last phase of the learning remains an issue, that has to be investigated in the future.
Further research also includes to experiment our Curriculum Learning procedure in combination with more elaborated optimization methods~\cite{Martens2011RNN_HF,Sutskever2013DNNMomentum}.

\subsection*{Acknowledgments}
This work was supported by
the French Research Agency under the contract Cognilego ANR 2010-CORD-013.

\ifNIPS
   \bibliographystyle{plain}
\else
   \bibliographystyle{IEEEtran}
\fi
\bibliography{biblio}

\end{document}